\documentclass[twocolumn]{article}
\usepackage{arxiv}

\usepackage[utf8]{inputenc}
\usepackage[english]{babel}
\usepackage{multicol}
\usepackage{float}
\usepackage{bbm}
\usepackage{amsmath,amsfonts}
\usepackage{multirow}
\usepackage{array}
\usepackage{graphicx}
\usepackage{booktabs}
\usepackage[numbers,sort&compress,square]{natbib}
\usepackage{xcolor}
\usepackage{tabularx}

\usepackage{blindtext}

\usepackage{amsmath}
\usepackage{autobreak}

\usepackage{enumitem}
\setlist[itemize]{noitemsep,topsep=3pt,leftmargin=*}

\usepackage[original]{abstract} %

\usepackage{balance}

\usepackage{hyperref}

\hypersetup{
pdftitle={SocialInteractionGAN: Multi-person Interaction Sequence Generation},
pdfsubject={Author version},
pdfauthor={Louis Airale, Dominique Vaufreydaz and~Xavier Alameda-Pineda},
pdfkeywords={Multi-person interactions, discrete sequence generation, adversarial learning},
}

\title{SocialInteractionGAN: Multi-person Interaction Sequence Generation}

\author{Louis~Airale$^{1,2}$, Dominique~Vaufreydaz$^1$ and~Xavier~Alameda-Pineda$^2$, IEEE Senior Member
\thanks{This research was partially funded in the framework of the ``Investissements d'avenir'' program (ANR-15-IDEX-02), by the H2020 SPRING (\#871245) and by the ANR ML3RI (ANR-19-CE33-0008-01).} \\
$^1$ Univ. Grenoble Alpes, CNRS, Grenoble INP, LIG, 38000 Grenoble, France\\
$^2$ Univ. Grenoble Alpes, Inria, CNRS, Grenoble INP, LJK, 38000 Grenoble, France
}

\begin{document}

\twocolumn[{%
  \begin{@twocolumnfalse}
    \maketitle
  \end{@twocolumnfalse}
  
}]

\saythanks 
\setcounter{footnote}{0}

        \begin{abstract}
Prediction of human actions in social interactions has important applications in the design of social robots or artificial avatars. In this paper, we focus on a unimodal representation of interactions and propose to tackle interaction generation in a data-driven fashion. In particular, we model human interaction generation as a discrete multi-sequence generation problem and present SocialInteractionGAN, a novel adversarial architecture for conditional interaction generation.
Our model builds on a recurrent encoder-decoder generator network and a dual-stream discriminator, that jointly evaluates the realism of interactions and individual action sequences and operates at different time scales.
Crucially, contextual information on interacting participants is shared among agents and reinjected in both the generation and the discriminator evaluation processes.
Experiments show that albeit dealing with low dimensional data, SocialInteractionGAN succeeds in producing high realism action sequences of interacting people, comparing favorably to a diversity of recurrent and convolutional discriminator baselines, and we argue that this work will constitute a first stone towards higher dimensional and multimodal interaction generation. Evaluations are conducted using classical GAN metrics, that we specifically adapt for discrete sequential data. Our model is shown to properly learn the dynamics of interaction sequences, while exploiting the full range of available actions.
    \end{abstract}
    

\vfill\break

\section{Introduction}

\begin{figure}[t]
\includegraphics[width=1\linewidth]{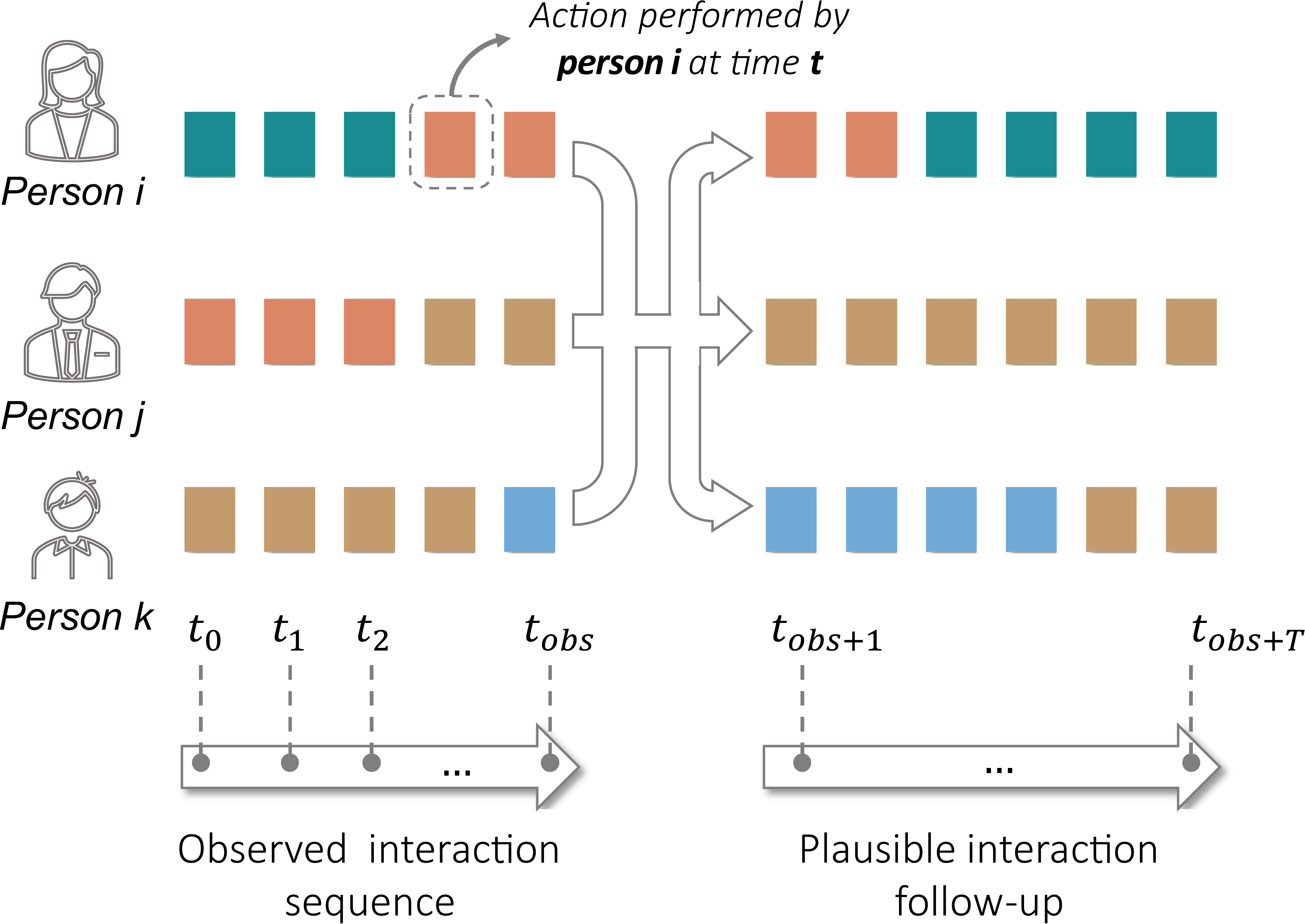}
\centering
\caption{Illustration of the task of interaction generation, where an agent performs one action at each discrete time step. Typically, the sampling frequency is chosen such that an action occurs over several time steps. Given an observed interaction sequence, a model should be able to generate realistic follow-ups for every participant.}
\label{fig:action_generation}
\end{figure}

Interactions between humans are the basis of social relationships, incorporating a large number of implicit and explicit multimodal signals expressed by the interaction partners~\cite{vinciarelli2009social}.
As the number of interacting people increases, so does the complexity of the underlying interpersonal synchrony patterns.
For humans, the interaction capability is at the same time innate and acquired through thousands of interaction experiences.
For interactive systems, predicting and generating human actions within social interactions is still far from the human-level performance. However, this task is central when it comes to devising, for instance, artificial avatars displaying realistic behavior or a social robot able to anticipate human intentions and adjust its response accordingly.
One possible explanation lies in the difficulty to collect and annotate human social behavioral data, illustrated by the scarcity of corpora available to the community.
A second source of difficulty is the intrinsically multimodal nature of human interactions. Ideally, when generating interaction data, a system should deal with signals as diverse as gaze, pose, facial expression or gestures.
In this work, we choose to relieve much of this complexity and treat interactions as synchronized sequences of "high level" discrete actions, and propose to solve the task of generating a realistic continuation to an observed interaction sequence (Figure~\ref{fig:action_generation}). The chosen representation has the advantage of easing the generation task while carrying enough meaningful information to study diverse interpersonal synchrony patterns.
This approach has in fact several benefits.
First, it is reasonable to think that part of the principles that govern discrete interactions will scale to other modalities, either discrete or continuous, as the same actions exist in the visual and audio domains. Therefore, the devised architecture may be extended to these further tasks.
Second, the generated interaction sequences will add up to the original dataset, representing as many fresh data samples. Provided a dataset that associates these simple actions with richer representations is available, one may use them as conditioning data for the generation of more complex interaction sequences.
Finally, representing the instantaneous state of a person in an interaction as a discrete action is consistent with the annotations provided in current human interaction datasets~\cite{cabrera2018matchnmingle,alameda2015salsa}.

The generation of discrete social action sequences of several people considering the interpersonal synchrony shares many ties with text generation in NLP and trajectory prediction in crowded environments. As in text generation, we seek to produce sequences of discrete data, or tokens, although the size of the token dictionary will typically be orders of magnitude smaller than that of any language. Discrete sequence generation in an adversarial setting comes with well identified limitations: non-differentiability that prevents backpropagating gradients from the discriminator, and lack of information on the underlying structure of the sequence~\cite{guo2018long}, due to the too restrictive nature of the binary signal of the discriminator. To overcome the first issue, the generator can be thought of as a stochastic policy trained with policy gradient~\cite{yu2017seqgan}. However, as noted in~\cite{guo2018long}, sequences produced via vanilla policy gradient methods see their quality decrease as their length grows, and therefore do not usually exceed 20 tokens. Another drawback of these methods is that they rely on several rounds of Monte Carlo search to estimate a policy reward, noticeably slowing down training.

As recent works in trajectory prediction exploit path information from surrounding persons to predict pedestrian trajectories that avoid collisions~\cite{alahi2016social,gupta2018social,lee2017desire}, one can think of interaction generation as the prediction of trajectories in a discrete action space where people constantly adapt their behaviour to other persons' reactions.
Because the action space is discrete, action generation is yet better modelled by sampling from a categorical distribution, which requires substantial changes from the previous works.
Moreover, and very differently from trajectories that are sequences of smoothly-changing continuous positions, discrete sequences of social actions cannot be considered as smooth. Accounting for other persons' behavior in social interactions therefore requires constant re-evaluations, because one must consider the recent actions performed by all the individuals within the same social interaction.
On the other hand, it should be possible to exploit the relative steadiness of action sequences: for a sampling rate of 25 fps, an action will typically extend for several tens of time steps. This fine grained resolution allows to deal with actions of different duration.

Last, this research faces a new challenge regarding the proper evaluation of 
the generated sequences of multi-person interactions. The Inception Score (IS) and the Fréchet Inception Distance (FID), two metrics commonly associated with adversarial generation, are primarily intended to assess image quality~\cite{salimans2016improved,heusel2017gans}. IS consists of two entropy calculations based on how generated images span over object classes, and it is relatively straightforward to adapt it to action sequences. FID, on the other hand, requires the use of a third party model usually trained on an independent image classification task. The absence of such an off-the-shelf inception model for discrete sequential data thus requires to devise a new independent task and to train the associated model that will serve to compute FID.

In this work, we present a conditional adversarial network for the generation of discrete action sequences of human interactions, able to produce high quality sequences of extended length even in the absence of supervision. 
We followed~\cite{alahi2016social,gupta2018social} and performed integration of contextual cues by means of a pooling module.
We used a discriminator with two distinct streams, one guiding the network into producing realistic action sequences, the other operating at the interaction level, assessing the realism of the participants’ actions relative to one another.
Noticeably, we built our model on a classical GAN framework as it demonstrated promising performances without the need of policy gradient. We propose however an essential local projection discriminator inspired from~\cite{miyato2018cgans} and~\cite{isola2017image} to provide the generator with localized assessments of sequence realism, therefore allowing the signal from the discriminator to be more informative.
The result is an adversarially trained network able to predict plausible future action sequences for a group of interacting people, conditioned on the observed beginning of the interaction.
Finally, we introduce two new metrics based on the principles of IS and FID so as to assess the quality and the diversity of generated sequences. In particular, we propose a general procedure to train an independent inception model on discrete sequences, whose intermediate activations can be used to compute a sequential FID (or SFID).

The contributions of this paper are:\vspace{-2mm}
\begin{itemize}
    \item A general framework for the generation of low dimensional multi-person interaction sequences;
    \item A novel dual-stream local recurrent discriminator architecture, that allows to efficiently assess the realism of the generated interaction sequences;
    \item Two variants of the popular Inception Score and Fréchet Inception Distance metrics, suited to assess the quality of sequence generation.
\end{itemize}

The rest of the manuscript is structured as follows. First, we review related work in trajectory prediction and text generation in section~\ref{section: related work}. We then describe our architecture in section~\ref{section: model}. In section~\ref{section metrics}, we introduce variants of Inception Score~\cite{salimans2016improved} and Fréchet Inception Distance~\cite{heusel2017gans}, widely used to assess the visual quality of GAN outputs, suited to our discrete sequence generation task. We then conduct experiments of the MatchNMingle dataset~\cite{cabrera2018matchnmingle} and show the superiority of our dual-stream local discriminator architecture over a variety of baseline models in section~\ref{section: experiments}. Finally, we conclude in section~\ref{section: conclusion}.

\section{Related work} \label{section: related work}
Following the above discussion, we split the description of the related work into two parts: trajectory prediction and generation of discrete sequential data.

\begin{figure*}
\includegraphics[width=1\linewidth]{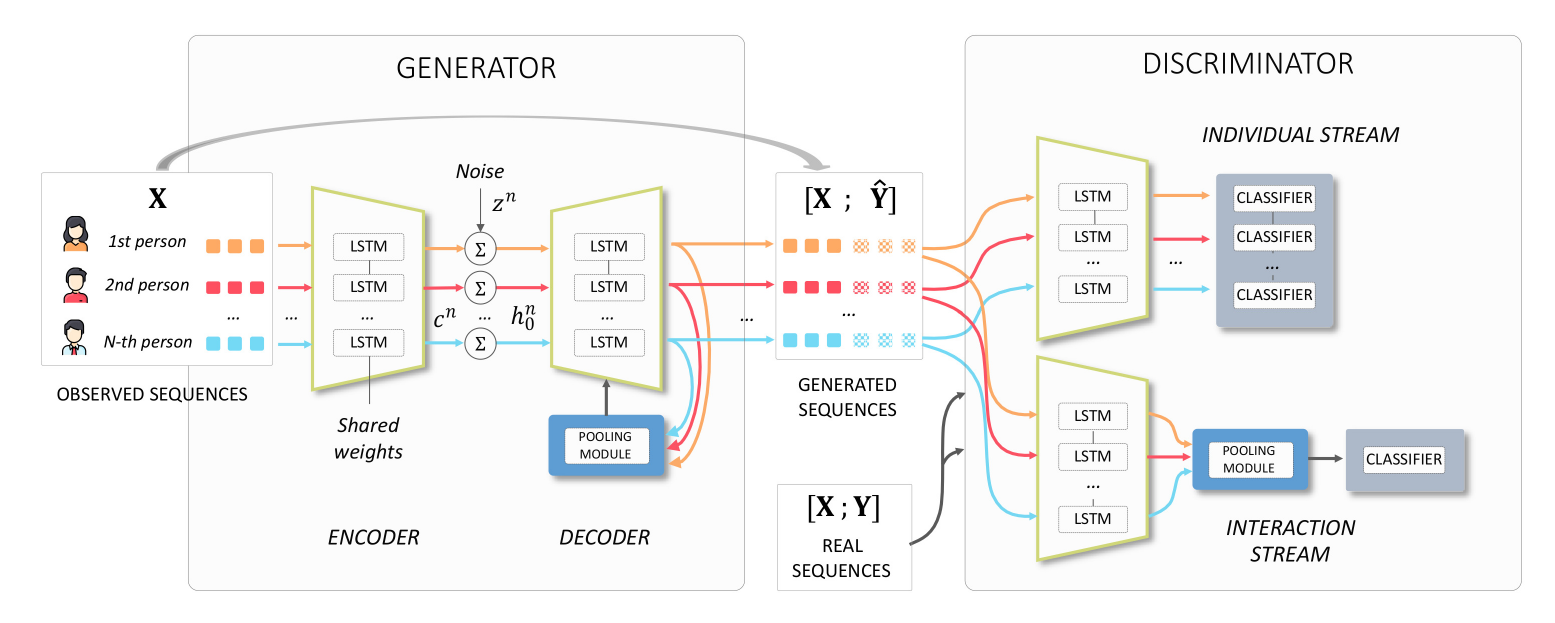}
\centering
\caption{Architecture of SocialInteractionGAN. Our model is composed of an encoder-decoder generator and a recurrent discriminator that assesses sequences individually and interaction as a whole. Pooling modules allow to mix hidden states from all participants to the interaction.}
\label{fig:architecture}
\end{figure*}

\subsection{Trajectory prediction}
Most recent works in trajectory prediction make use of contextual information of surrounding moving agents or static scene elements to infer a trajectory~\cite{alahi2016social,fernando2018soft+,bartoli2018context,gupta2018social,lee2017desire,zhao2019multi,sadeghian2019sophie,vasishta:hal-01875147}.
In~\cite{lee2017desire}, a conditional VAE is used to generate a set of possible trajectories, which are then iteratively refined by the adjunction of cues from the scene and surrounding pedestrians. In~\cite{sadeghian2019sophie} the authors propose a GAN and two attention mechanisms to select the most relevant information among spatial and human contextual cues for the considered agent path. As in previous works, the authors of~\cite{zhao2019multi} use an encoder-decoder architecture, but propose to jointly exploit pedestrian and static scene context vectors using a convolutional network. The resulting vectors are then spacially added to the output of the encoder before the decoding process. In the preceding works, pooling strategies are usually employed to enforce invariance to the number of participants in the scene. Such strategies include local sum pooling~\cite{alahi2016social}, average pooling~\cite{lee2017desire}, or more sophisticated procedures involving non-linear transformations~\cite{gupta2018social}. 
In our setting, we posit an equal prior influence of all interacting participants on the decision to perform an action, and leave the exploration of different conditions to future work. This alleviates the need for a pooling strategy aware of the spatial position. However it is important that the contextual information provided to the decoder should follow the course of the interaction, and thus be recomputed at every time step. This is similar to~\cite{fernando2018soft+,sadeghian2019sophie} that rely on a time-varying attention context vector.

In most previous works the models are either trained to minimize the mean square distance from a ground truth trajectory or to maximize the realism of single person trajectories thanks to an adversarial loss, but few consider trajectories interplay in the training objective. This is the case of SocialGAN~\cite{gupta2018social}, where the discriminator outputs a single score for the entire scene, while the adequacy of individual trajectories is ensured by a supervised (\(L_2\)) loss and by the prediction of small incremental displacements over time steps. In contrary, we would like our adversarial loss to cover both the realism of individual sequences and of the interaction as a whole, and we achieve this by using a discriminator with two streams.
This way we are able to investigate the effects of lessening the weight of the supervised \(L_2\) loss or to simply train the model without it.
Note that we could follow a similar strategy to~\cite{zhao2019multi} and sum individual hidden states with contextual information into a single vector serving as input to the discriminator classifier. We want however to preserve as much information as possible from the two streams and maintain two separate classifiers.

\subsection{Generation of discrete sequential data}
GANs have recently been proposed to complement the traditional Maximum Likelihood Estimation training for text generation to mitigate the so-called exposure bias issue, with pretty good results on short word sequences~\cite{li-etal-2017-adversarial,yu2017seqgan,lin2017adversarial}. The use of an adversarial loss for discrete sequence generation network yet comes with two well-identified caveats. First, workarounds must be found to allow the architecture to be fully differentiable. Second, the binary signal from the discriminator is scarce and may lack information as for instance for what in a sentence makes it unrealistic~\cite{guo2018long}. In~\cite{yu2017seqgan} and~\cite{li-etal-2017-adversarial} authors interpret text generation as a Markov Decision Process where an agent selects a word at each state, constituted of all previously generated words. The generator is trained via stochastic policy gradient, while the reward is provided by the discriminator. These ideas are extended in~\cite{lin2017adversarial}, where the discriminator is modelled as a ranker rather than a binary classifier to avoid vanishing gradient issues. In~\cite{guo2018long}, the generator is implemented as a hierarchical agent, following the architecture proposed in~\cite{vezhnevets2017feudal}, so as to produce text sequences of increased length. Action sequences however differ from text by the limited size of action space compared to a language dictionary. We also hypothesize a shorter ``memory'' of action sequences: it is likely that one can judge the overall quality of an action sequence by looking only at small chunks of it, provided their length is chosen adequately. Therefore we propose a local discriminator operating on short-range receptive fields to allow its signal to be more informative, which can be seen as a recurrent equivalent of PatchGAN~\cite{isola2017image}. Plus, we found that using a classical adversarial loss was sufficient to achieve high quality results, which relieved us with the burden of keeping an estimate of the expected reward.

\section{Multi-person interaction sequence generation with SocialInteractionGAN}
\label{section: model}

We present SocialInteractionGAN, a conditional GAN for multi-person interaction sequence generation. Our model takes as input an interaction \(\mathbf{X}\) of duration \(t_{obs}\), which is constituted of \(N\) synchronized action sequences, one for each participant. We denote the observed action sequence of person \(n\) as \(\mathbf{x}^n_{1:t_{obs}} = \{x_\tau^n, 1 \leq \tau \leq t_{obs}\}\) (in practice we will drop time index to simplify notations and only use \(\mathbf{x}^n\) for person \(n\) input as it is always taken between \(\tau=1\) and \(\tau=t_{obs}\)). Our goal is to predict socially plausible future actions for all interacting agents. We write the newly generated interaction as \(\hat{\mathbf{Y}} = (\hat{\mathbf{y}}_{1:T}^1,\ldots, \hat{\mathbf{y}}_{1:T}^N)\), where similarly \(\hat{\mathbf{y}}_{1:T}^n = \{\hat{y}_\tau^n, t_{obs} + 1 \leq \tau \leq t_{obs} + T\}\) is the generated action sequence for person \(n\). Again \(\hat{\mathbf{y}}^n\) will be employed when referring to the whole generated sequence.
The generated interaction \(\hat{\mathbf{Y}}\) can be compared with the ground truth \(\mathbf{Y}\).

\subsection{SocialInteractionGAN architecture overview}
As illustrated in Figure~\ref{fig:architecture}, our model is composed of a recurrent encoder-decoder generator network (see section~\ref{subsection: generator}) and a dual-stream local recurrent discriminator (see section~\ref{subsection: D}). The generator encodes the observed part of the interaction and sequentially generates future actions for all participants. We use a pooling module to merge all participants' hidden states into a single vector that is fed to the decoder at each time step. This way, the decoded actions are obtained using a compact representation of the previous actions of all surrounding persons. The pooling operation is invariant to the number of interacting people, and so is our model that can therefore virtually handle interacting groups of any size.
The generated interaction \(\hat{\mathbf{Y}}\) is concatenated to the conditioning tensor \(\mathbf{X}\) and input to the discriminator, alternatively with the actual interaction \([\mathbf{X}, \mathbf{Y}]\). Our dual-stream discriminator then assesses the realism of both individual action sequences and interaction as a whole.
We train the network using these two adversarial losses and ablate the use of a supervised \(L_2\) loss. The following sections detail the various modules of the architecture in Figure~\ref{fig:architecture}.

\subsection{Generator}
\label{subsection: generator}
\subsubsection{Encoder Network}

Our encoder is based on a recurrent network that operates on a dense embedding space of the discrete actions. The input sequences are processed by a LSTM recurrent encoder \(f_e\)~\cite{hochreiter1997long}, independently for each person \(n\):
\begin{equation}
    c^n = f_e(\mathbf{x}^n).
\end{equation}

A random noise vector \(z^n\) of the same dimension as \(c^n\) is then added to the encoding of the observed sequence. This is to allow the GAN model to generate diverse output sequences, as the problem of future action generation is by essence non-deterministic:
\begin{equation}
    h_0^n = c^n + z^n.
\end{equation} 
The resulting vector \(h_0^n\) is then used to initialize the decoder's hidden state.

\subsubsection{Decoder Network}
The decoder is also a recurrent network that accounts for the actions of all persons involved in the interaction thanks to a pooling module. This pooling  module, which is essentially a max operator, is responsible for injecting contextual cues in the decoding process, so that the generator makes an informed prediction when selecting the next action. Such a pooling is essential to let the number of interacting people vary freely. Concretely, at each time step \(\tau\) the decoder takes as input the preceding hidden state \(h_{\tau-1}^n\) and decoded action \(\hat{y}_{\tau-1}^n\), and a vector \(h_{\tau-1}\) output by the pooling module (which is person-independent). A deep output transformation \(g\), implemented as a multi-layer perceptron, is then applied on the resulting hidden state \(h_{\tau}^n\). One ends up with action probabilities \(p(y_{\tau}^n)\), in a very classical sequence transduction procedure, see e.g.~\cite{bahdanau2014neural}.  Formally, we have:
\begin{align}
    h_{\tau}^n &= f_d(h_{\tau-1}^n; \hat{y}_{\tau-1}^n, h_{\tau-1})\\
    p(y_{\tau}^n) &= g(h_{\tau}^n, \hat{y}_{\tau-1}^n, h_{\tau-1}),
\end{align}
where \(f_d\) is the decoder recurrent network, also implemented as a LSTM. 

The \(j\)-th coordinate of the output of the pooling writes:
\begin{equation}
    h_{\tau,j} = \underset{n}{\max} \: h_{\tau, j}^n
\end{equation}
where \(h_{\tau, j}^n\) designates the \(j\)-th coordinate of person \(n\) hidden state at time \(\tau\).

The resulting action \(\hat{y}_{\tau}^n\) then needs to be sampled from \(p(y_{\tau}^n)\). As we want our discriminator to operate in the discrete action space, we use a softmax function with temperature \(P\) as a differentiable proxy when sampling for the discriminator, i.e. the \(\tau\)-th entry of discriminator input sequence for person \(n\) writes:
\begin{equation}
    \hat{y}_{\tau}^n = \text{softmax}(p(y_{\tau}^n)/P)
\end{equation}
where \(P\) is typically equal to \(0.1\), i.e. small enough so that softmax output is close to a one-hot vector.

\subsection{Discriminator} \label{subsection: D}

The discriminator can be implemented in many ways, but it usually relies on recurrent (RNN) or convolutional (CNN) networks, depending on the application.
Trajectory prediction applications usually rely on recurrent networks~\cite{gupta2018social,sadeghian2019sophie, zhao2019multi}, whereas convolutional networks are preferred for text generation~\cite{yu2017seqgan,lin2017adversarial,guo2018long}, as CNN performances were shown to surpass that of recurrent networks on many NLP tasks~\cite{zhang2015text,gehring2017convolutional}. Convolutional networks also have the advantage to be inherently suitable to parallel computing, drastically reducing their computation time. Borrowing from both paradigms, we turn to a recurrent architecture that lends itself to batch computing, while preserving temporal progression in sequence analysis.

\subsubsection{Dual-stream discriminator}

Two streams coexist within our discriminator such that the realism of both action sequences and participants interactions can be explicitly enforced. The individual stream (see Figure~\ref{fig:architecture}), labelled as \(D_{indiv}\), is composed of a recurrent network followed by a classifier, respectively implemented as a LSTM and a shallow feed-forward neural network, whose architecture is detailed in the following section. It operates on individual action sequences, assessing their intrinsic realism disregarding any contextual information. The interaction stream, \(D_{inter}\), follows the same architectural lines, but a pooling module similar to the one used in the generator is added right after the recurrent network such that the classifier takes a single input for the whole interaction. A factor \(\lambda_{inter}\) controls the relative importance of interaction stream, such that the full discriminator writes:
\begin{equation}
    D_{tot} = D_{indiv} + \lambda_{inter} D_{inter}.
\end{equation}

\subsubsection{Local projection discriminator}

Implementing (any of) the two discriminators as a recurrent network and letting them assess the quality of the entire generated sequences poses several issues. First the contributions from different sequence time steps to weight updates is likely to be unbalanced due to possible vanishing or exploding gradients.
Second, some of those gradients may be uninformative especially at the onset of training when errors are propagated forward in the generation process, degrading the quality of the whole sequence, or when the overall realism depends on localized patterns in the data. On the contrary we seek for a discriminator architecture that is better able to guide the generation with gradients corresponding to local evaluations, so as to entail an even contribution from each location in the generated sequence. To that end, previous works used a CNN discriminator~\cite{yu2017seqgan,lin2017adversarial}. We explore recurrent architectures that conform with that objective. Along the same lines as PatchGAN~\cite{isola2017image} where photorealism is computed locally at different resolutions, we propose a multi-timescale local discriminator, applying on overlapping sequence chunks of increasing width. Another intuition drove us to this choice: it seems reasonable to assume that the realism of an action sequence can be assessed locally, while it would not be the case for text sequences where preserving meaning or tense throughout a sentence appears more crucial. To that aim, the generated action sequence \(\hat{\mathbf{y}}^n\) is split into \(K\) overlapping subsequences temporally indexed by \(\tau_1, \ldots, \tau_K\), with \(K\) uniquely defined by the chunk length \(\Delta \tau\) and the interval \(\delta \tau = \tau_{k+1} - \tau_k\) between successive chunks (see section~\ref{section: experiments} for a detailed discussion on how to select \(\Delta \tau\) and \(\delta \tau\)).
Each subsequence is then processed independently through the recurrent module \(f\) of the discriminator:
\begin{equation}
    h_{\tau_k}^n = f(\hat{\mathbf{y}}_{\tau_k: \tau_k + \Delta \tau}^n),
\end{equation}
\(\hat{\mathbf{y}}_{\tau_k: \tau_k + \Delta \tau}^n\) being the \(k\)-th action subsequence of person \(n\) (hence comprised between time steps \(\tau_k\) and \(\tau_k + \Delta \tau\), plus the offset \(t_{obs}\)). Next, to account for the conditioning sequence \(\mathbf{x}^n\) and its resulting code \(h^n = f(\mathbf{x}^n)\), we implement a projection discriminator~\cite{miyato2018cgans} and dampen the conditioning effect as we move away from the initial sequence thanks to  a trainable attenuation coefficient.
Discriminator output can finally be written as follows:
\begin{equation} \label{eq:D}
    D(\mathbf{x}^n, \hat{\mathbf{y}}^n) = \frac{1}{K} \sum_k D_{proj}(h^n, h_{\tau_k}^n),
\end{equation}
with
\begin{multline}
    D_{proj}(h^n, h_{\tau_k}^n) = A(\tau_k^{-1/\beta} (h^n)^\top V \phi(h_{\tau_k}^n) \mathbb{1}_{d_\psi} \\ + \psi(\phi(h_{\tau_k}^n))),
\end{multline}
where \(\mathbb{1}_{d_\psi}\) is the vector of ones of size \(d_\psi\), \(\phi\) and \(\psi\) are fully-connected layers, \(A\) and \(V\) real-value matrices whose weights are learned along other discriminator parameters, and \(\beta\) a trainable parameter that controls the conditioning attenuation. In our case, given \(h^n\), \(h_{\tau_k}^n \in \mathbb{R}^{d_h}\), \(\phi(\cdot) \in \mathbb{R}^{d_\phi}\), \(\psi(\cdot) \in \mathbb{R}^{d_\psi}\), then \(V \in \mathbb{R}^{d_h \times d_\phi}\) and \(A \in \mathbb{R}^{1 \times d_\psi}\). 

We repeat the same procedure over different chunk sizes \(\Delta \tau\) to account for larger or smaller scale patterns, and average the scores to give the final output from the individual stream \(D_{indiv}\). Slight differences arise for the computation of \(D_{inter}\). Summation terms in (\ref{eq:D}) are no longer evaluated independently for each participant. Instead, the \(N\) vectors \(h_{\tau_k}^1\), \ldots, \(h_{\tau_k}^N\) output by the encoder for all \(N\) participants at time \(\tau_k\) are first processed through the pooling module, yielding the pooled vector \(h_{\tau_k}\). We do the same for individual conditioning vectors \(h^n\), that are pooled to give a single conditioning vector \(h\) for the whole interaction. \(D_{proj}\) is then evaluated on \(h\) and \(h_{\tau_k}\) in (\ref{eq:D}), and its output indicates how much the chunk of interaction starting at \(\tau_k\) is realistic given the observed interaction \(\mathbf{X}\).

\subsection{SocialInteractionGAN training losses}

We use an adversarial hinge loss~\cite{lim2017geometric} to train our model, i.e. for the discriminator the loss writes:

\begin{multline}
    \mathcal{L}_D = \mathbb{E}_{\mathbf{X}} \mathbb{E}_{\hat{\mathbf{Y}}} [\max(0, 1 + D_{tot}(\mathbf{X}, \hat{\mathbf{Y}})) \\ + \max(0, 1 - D_{tot}(\mathbf{X}, \mathbf{Y}))],
\end{multline}
as for the generator, the adversarial loss writes:

\begin{equation}
    \mathcal{L}_G^{adv} = -\mathbb{E}_{\mathbf{X}} \mathbb{E}_{\hat{\mathbf{Y}}} [D_{tot}(\mathbf{X},  \hat{\mathbf{Y}})]
\end{equation}
where the expectation is taken over dataset interactions \(\mathbf{X}\) and the random matrix of model outputs \(\hat{\mathbf{Y}}\), with \(\mathbf{Y}\) the ground truth sequence.

We also add the following supervision loss to the generator:

\begin{equation}
    \mathcal{L}_G^{sup} = \mathbb{E}_{\mathbf{X}} \mathbb{E}_{\hat{\mathbf{Y}}} \left[ \|\hat{\mathbf{Y}} - \mathbf{Y}\|_2^2 \right]
\end{equation}
(where \(\hat{\mathbf{Y}}\) and \(\mathbf{Y}\) dependency on \(\mathbf{X}\) is implicit) and weight it with a factor \(\lambda_{sup}\). This last factor must be carefully chosen such that training mainly relies on the adversarial loss. This is important because, as showed in the experiments, the mean squared error fails to explore the diversity of action sequences. However, a small values of \(\lambda_{sup}\) can have an interesting stabilizing effect on training (see section~\ref{section: experiments}). The overall generator loss writes:

\begin{equation}
    \mathcal{L}_G = \mathcal{L}_G^{adv} + \lambda_{sup} \mathcal{L}_G^{sup}.
\end{equation}

\section{Evaluation of generated sequences quality}
\label{section metrics}

Inception Score (IS) and Fréchet Inception Distance (FID) are two metrics designed to assess the quality and diversity of GAN-generated images and compare the distributions of synthetic and real data~\cite{salimans2016improved,heusel2017gans}. In the absence of sequence class labels for computing IS and of an auxiliary inception model for FID, neither of these two metrics can however directly apply to discrete sequences.
We begin by re-writing the calculation of IS.
The same original formula is used:

\begin{equation}
    IS = \exp(H_M - H_C)
\end{equation}
with \(H_M\) and \(H_C\) respectively the marginal and conditional entropies, but we rely on action labels to estimate the sequential equivalents of those two variables. The marginal entropy is computed over the entire set of predicted sequences and measures how well the GAN learned to generate diverse actions. It is therefore expected to be high. As for the conditional entropy, it needs to be low in a model that learned to grasp key features of a class, and thus we define it as the average entropy of individual sequences. Formally, we write: 
\begin{align}
    H_M &= - \sum_{a\in \mathcal{A}} f(a)\log(f(a)) \\
    H_C &= - \frac{1}{|\mathcal{S}|} \sum_{a\in \mathcal{A}, s \in \mathcal{S}} f_s(a)\log(f_s(a))
\end{align}
where \(\mathcal{A}\) is the set of actions, \(f(a)\) the frequency of action \(a\in \mathcal{A}\) over all generated sequences, and \(f_s(a)\) the frequency of action \(a\) in sequence \(s\) of the dataset \(\mathcal{S}\) of size \(|\mathcal{S}|\).

This new definition of \(H_C\) suffers from a limitation. Indeed, the conditional entropy decreases as the model learns to generate steadier and consistent sequences. However, very low values of \(H_C\) means that the same action is repeated over the whole sequence.
Rather that aiming for the lowest possible values of \(IS\), we therefore compute \(H_M\) and \(H_C\) from the data and use them as oracle values. In addition, we complement these scores with a sequential equivalent of FID so as to compare the distributions of real and generated sequences.

FID measures the distance between two image distributions by comparing the expectations and covariance matrices of intermediate activations of an Inception v3 network~\cite{szegedy2016rethinking} trained on image classification~\cite{heusel2017gans}.
However Iv3 is irrelevant for our discrete sequential data.
Therefore we built a recurrent inception network that we trained on an auxiliary task. 
We implemented it as a bidirectional LSTM encoder, followed by five feed-forward layers, and trained it to regress the proportion of each action in given input sequences.
The main principle that led to the choice of this task was that it could be used to finely characterize the input sequences, ensuring hidden activations to contain meaningful representations of sequences. In particular, it seems plausible to assume that the realism of an action sequence can be partly assessed simply knowing the proportion of each action.
Finally, FID distances were computed by extracting the last activations before the regression head.
We refer to the distances computed with this proposed metric as SFID for Sequential Fréchet Inception Distance.

\section{Experiments} \label{section: experiments}

We conducted our experiments on the MatchNMingle dataset~\cite{cabrera2018matchnmingle}, that contains action-annotated multi-person interaction data that are particularly suited to our task. 
Although other datasets featuring human interaction data exist (e.g.~\cite{mccowan2005ami,alameda2015salsa,KOUTSOMBOGERA18.596,kossaifi2019sewa}), they do not fit in the framework described here, either because the discrete annotations are not exhaustive, or because they rely on transcripts that are not readily usable or on continuous quantities like body pose or raw video that are out of the scope of the present study.
On the contrary, the MatchNMingle dataset contains annotated actions that fully define the state of all participants at each time step.

We carried out pre-processing on the original data, that we detail in  section~\ref{subsection: pre-processing}. In the absence of concurrent work, we challenged our architectural choices versus alternative recurrent and convolutional discriminator baselines (sections~\ref{section alternative D} and~\ref{section cnn}) and carried an ablation study~(section~\ref{section ablation}), highlighting the relevance of our dual-stream local discriminator.

\subsection{MatchNMingle dataset}
The MatchNMingle dataset contains annotated video recordings of social interactions of two kinds: face-to-face speed dating (``Match'') and cocktail party (``Mingle''), out of which we exclusively focused on the latter. Mingle data contains frame-by-frame action annotated data for a duration of 10 minutes on three consecutive days, each of them gathering  approximately 30 different people. Each frame is decomposed into small groups of chatting people that constitute our independent data samples.
Mingle data comprises for instance 33 interactions of three people that amount to a total of 51 minutes of recording, annotated at a rate of 20 frames per second. We focused our experiments on three-people interactions as it offers a good trade-off between a sufficient complexity of action synchrony patterns and generalization capability that is dependent of the quantity of available data.
Interactions are composed of eight different labelled actions: \textit{walking}, \textit{stepping}, \textit{drinking}, \textit{hand} \& \textit{head gesture}, \textit{hair touching}, \textit{speaking} and \textit{laughing}, plus an indicator of occlusion that can be total or partial.

\begin{table}
\centering
\caption{Comparison of recurrent discriminator baselines for sequences of 40 time steps, with weak supervision (above) and no supervision (below). Marginal entropy (\(\text{H}_\text{M}\)), conditional entropy (\(\text{H}_\text{C}\)) and SFID correspond to the training epoch that yields the best results. Best models are those that achieve the closest \(\text{H}_\text{M}\) and \(\text{H}_\text{C}\) values to the real data and the lowest SFID.}
\label{table:1}
\resizebox{.5\textwidth}{!}{
\begin{tabular}{m{5em} m{3em} m{5.2em} m{5.2em} m{5.2em}} 
\toprule
Model & \(\lambda_{sup}\) & \(\text{H}_\text{M}\) & \(\text{H}_\text{C}\)&  SFID\\[0.5ex] 
 \midrule
 \multicolumn{2}{l}{\textit{Real Data}}&\(\textit{2.18}\)&\(\textit{0.30}\)& --\\
 \midrule
 SimpleRNN & \(10^{-3}\) & \(2.07 \pm 0.06\) & \(0.04 \pm 0.03\) & \(0.72 \pm 0.22\)\\
 DenseRNN & \(10^{-3}\) & \(2.15 \pm 0.05\)  & \(0.04 \pm 0.02\) &  \(0.84 \pm 0.21\)\\
 LocalRNN & \(10^{-3}\) & \(\textbf{2.18} \pm \textbf{0.04}\) & \(\textbf{0.26} \pm \textbf{0.06}\)  &  \(\textbf{0.41} \pm \textbf{0.09}\)\\
 \midrule
 SimpleRNN & \(0\) & \(1.94 \pm 0.05\) & \(0.38 \pm 0.07\)  & \(1.10 \pm 0.25\)\\
 DenseRNN & \(0\) & \(2.20\pm 0.07\) & \(0.22 \pm 0.05\) &  \(0.44 \pm 0.09\)\\
 LocalRNN & \(0\) & \(\textbf{2.18} \pm \textbf{0.03}\)  & \(\textbf{0.26} \pm \textbf{0.03}\)  &  \(\textbf{0.24} \pm \textbf{0.04}\)\\
 \bottomrule
\end{tabular}}
\end{table}

\begin{table}
\centering
\caption{Comparison of recurrent discriminator baselines for sequences of 80 time steps, with weak supervision (above) and no supervision (below). Best models are those that achieve the closest \(\text{H}_\text{M}\) and \(\text{H}_\text{C}\) values to the real data and the lowest SFID.}
\label{table:2}
\resizebox{.5\textwidth}{!}{
\begin{tabular}{m{5em} m{3em} m{5.2em} m{5.2em} m{5.2em}} 
\toprule
Model & \(\lambda_{sup}\) & \(\text{H}_\text{M}\) & \(\text{H}_\text{C}\)& SFID\\[0.5ex] 
\midrule
 \multicolumn{2}{l}{\textit{Real Data}}&\(\textit{2.18}\)&\(\textit{0.51}\)& --\\
\midrule
 SimpleRNN & \(10^{-3}\) & \(1.98 \pm 0.06\)  &  \(0.26 \pm 0.13\)  & \(1.41 \pm 0.36\)\\
 DenseRNN & \(10^{-3}\) & \(2.12 \pm 0.03\)  & \(0.05 \pm 0.04\) &  \(0.82 \pm 0.06\)\\
 LocalRNN & \(10^{-3}\) & \(\textbf{2.16} \pm \textbf{0.05}\)  & \(\textbf{0.34} \pm \textbf{0.09}\)  &  \(\textbf{0.74} \pm \textbf{0.27}\)\\
 \midrule
 SimpleRNN & \(0\) & \(1.96 \pm 0.03\)  &  \(0.26 \pm 0.08\)  & \(1.87 \pm 0.36\)\\
 DenseRNN & \(0\) & \(2.13 \pm 0.13\)  & \(0.29 \pm 0.15\) &  \(1.40 \pm 0.53\)\\
 LocalRNN & \(0\) & \(\textbf{2.15} \pm \textbf{0.06}\)  & \(\textbf{0.45} \pm \textbf{0.10}\)  &  \(\textbf{0.73} \pm \textbf{0.26}\)\\
 \bottomrule
 
\end{tabular}}
\end{table}

\begin{figure*}
\includegraphics[width=1\linewidth]{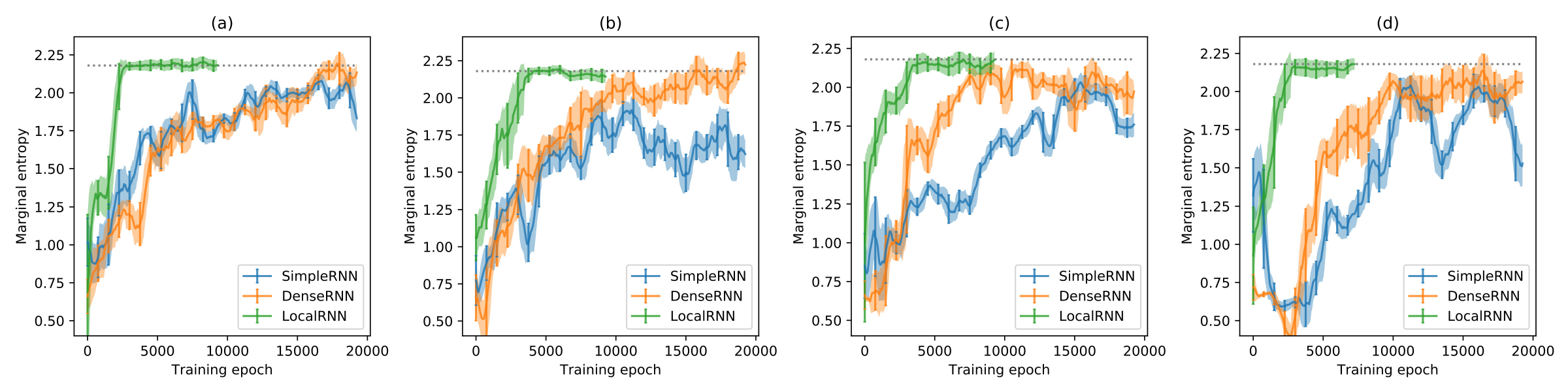}
\centering
\caption{Evolution of marginal entropy for different recurrent discriminator architectures and training configurations: (a) sequences of 40 time steps with \(\lambda_{sup}=10^{-3}\); (b) sequences of 40 time steps with \(\lambda_{sup}=0\); (c) sequences of 80 time steps with \(\lambda_{sup}=10^{-3}\); (d) sequences of 80 time steps with \(\lambda_{sup}=0\). Gray dotted line represents the marginal entropy of the data (equal to 2.18). LocalRNN converges in all cases to data-like marginal entropy values, even for long sequences and unsupervised training (hence the early stopping), which is not the case for the two other baselines.}
\label{fig:hm_all_rnn}
\end{figure*}

\subsection{Data pre-processing}
\label{subsection: pre-processing}
To build our dataset we processed Mingle data as follows. We first considered each group of interacting people independently and split the interactions into non-overlapping segments of three seconds (60 frames). These 3-second segments constituted the conditioning data~\(\mathbf{X}\).
We considered each segment's following actions as the target sequences~\(\mathbf{Y}\). Out of this dataset, we removed all training samples in which total occlusion accounted for more than ten percent of the sample actions, so as to limit the impact of the occlusions in the dataset.
Our dataset finally comprises 600 three-people interaction samples.
Finally, in order to ease data manipulation, we built an action dictionary containing the most common actions or combinations of actions, and replaced the 8-dimensional binary action vectors of the data by one-hot vectors drawn from the dictionary. We let the cumulative occurrences of dictionary entries amount to 90\% of observed actions to prevent the model from struggling with rare action combinations, and gathered all remaining actions under an additional common label. Experiments with actions accumulating up to 99\% of occurrences are reported in section~\ref{section adverse}. The resulting dictionary contains the 14 following entries: \textit{no action}, \textit{speaking + hand gesture}, \textit{speaking}, \textit{stepping}, \textit{head gesture}, \textit{hand gesture}, \textit{drinking}, \textit{speaking + hand gesture + head gesture}, \textit{hand gesture + head gesture}, \textit{speaking + head gesture}, \textit{stepping + speaking + hand gesture + head gesture}, \textit{stepping + hand gesture + head gesture}, \textit{stepping + hand gesture}, \textit{laughing}.

\subsection{Experimental details} \label{section: exp_details}
In all our experiments we used layer normalization~\cite{ba2016layer} in all recurrent networks, including that of SFID inception network, for its stabilization effect on the gradients, along with spectral normalization~\cite{miyato2018spectral} after each linear layer and batch normalization~\cite{ioffe2015batch} in decoder deep output. All recurrent cells were implemented as LSTMs~\cite{hochreiter1997long} with a hidden state dimension \(d_h = 64\), and we chose the same dimension for the embedding space. The dimensions of projection discriminator dense layers were chosen to be \(d_\phi = d_\psi = 128\). For the default configuration of our local discriminator, we used four different chunk sizes \(\Delta \tau\). Three of them depend on the output sequence length \(T\): \(T\), \(T/2\), \(T/4\) and we fixed the smallest chunk size to \(5\). We set \(\delta\tau = \Delta \tau / 2\), thus ensuring 50\% overlap (rounded down) between consecutive chunks for all four resolutions. Different configurations are explored in section~\ref{section cnn}. For comparison purpose we built the CNN baselines in a similar fashion, with parallel streams operating at different resolutions and kernels of the same sizes as the chunk widths defined above.
In all experiments, we use Adam optimizer~\cite{Kingma2015AdamAM} and set learning to \(2.10^{-5}\) for the generator and \(1.10^{-5}\) for the discriminator. As for the training of SFID inception net, we set the learning rate to \(1.10^{-3}\).

\subsection{Experimental results}

We articulate our experiments as follows: a comparison of recurrent model baselines, a comparison of recurrent and convolutional discriminators and an ablation study, that support the different architectural choices of our network.

\subsubsection{Alternative recurrent discriminator baselines} \label{section alternative D}

We start by comparing several recurrent architecture baselines for the discriminator, under different supervision conditions and for 40 and 80 frame-long sequences (respectively 2 and 4 seconds). In section~\ref{subsection: D}, we motivated our architectural choices on the hypothesis that a generator would benefit preferentially from multiple local evaluations rather that fewer ones carried on larger time ranges. To support this assumption, we evaluated our local discriminator (hereafter LocalRNN) against two baselines: the first one, labelled as SimpleRNN, only processes the whole sequence at once and outputs a single realism score. The second one, referred to as DenseRNN, also processes the sequence at once, but in this case all intermediate hidden vectors are conserved and used as input for the classifier. In this way, the discriminator output contains also localized information about actions, although the contributions to the score and the gradients of different time steps remain unbalanced. Results are gathered in Tables~\ref{table:1} and~\ref{table:2} for sequences of 40 and 80 time steps respectively, and correspond to the epoch that yields the best results for each model in terms of marginal entropy and SFID (longer training sometimes results in degraded performance). Corresponding marginal entropy evolutions are displayed in Figure~\ref{fig:hm_all_rnn}. LocalRNN consistently produces sequences of data-like quality in terms of marginal and conditional entropies, regardless of sequence length or supervision strength, and achieves the lowest SFID scores.
Interestingly, we notice that if SimpleRNN seems to benefit from a weak \(L_2\) as is suggested by a lower SFID, it is not the case for the two local models. The supervised loss has a squishing effect on the conditional entropy that is particularly detrimental for DenseRNN, yielding a model that does not properly generate action transitions. Finally, as one can see from the training dynamics (Figure~\ref{fig:hm_all_rnn}), all LocalRNN models converge within three thousands training epochs, a much shorter time than any of the other recurrent discriminator baselines. Last but not least, the long-range recurrent evaluations of Simple and DenseRNN discriminators (on sequences of length \(t_{obs} + T\)) are replaced in LocalRNN by many short-range ones that can be efficiently batched, resulting in a much shorter inference time. This fast and efficient training advocates in favor of our local discriminator over recurrent architectures with larger focal.

\begin{figure*}
\includegraphics[width=1\linewidth]{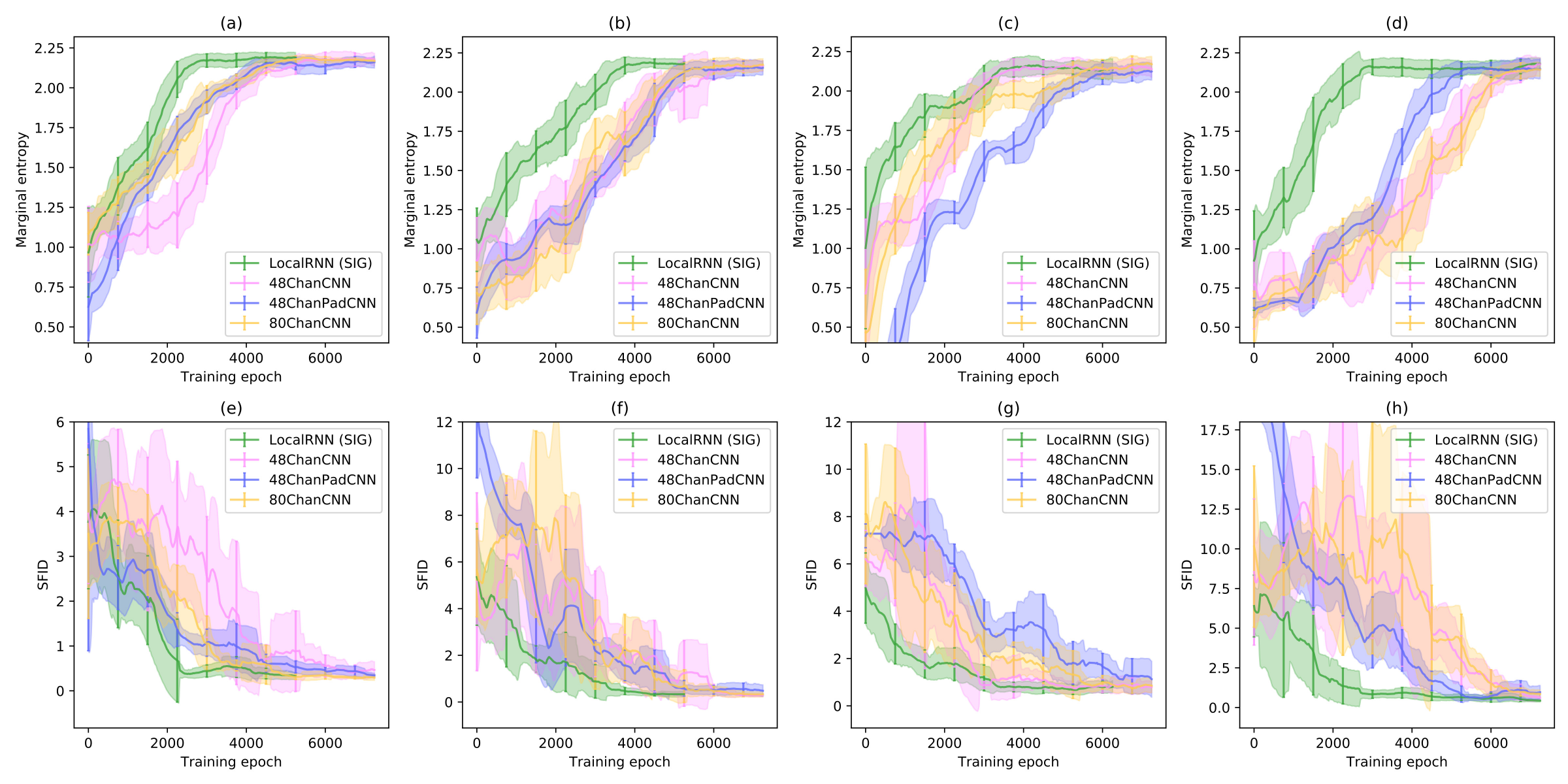}
\centering
\caption{Evolution of marginal entropy and SFID of SocialInteractionGAN (LocalRNN) and various CNN discriminator baselines, for the following training configurations: (a) sequences of 40 time steps with \(\lambda_{sup}=10^{-3}\); (b) sequences of 40 time steps with \(\lambda_{sup}=0\); (c) sequences of 80 time steps with \(\lambda_{sup}=10^{-3}\); (d) sequences of 80 time steps with \(\lambda_{sup}=0\).}
\label{fig:hm_cnn_rnn}
\end{figure*}

\subsubsection{Local recurrent vs. convolutional discriminators} \label{section cnn}

\begin{figure}
\centering
\includegraphics[width=1\linewidth]{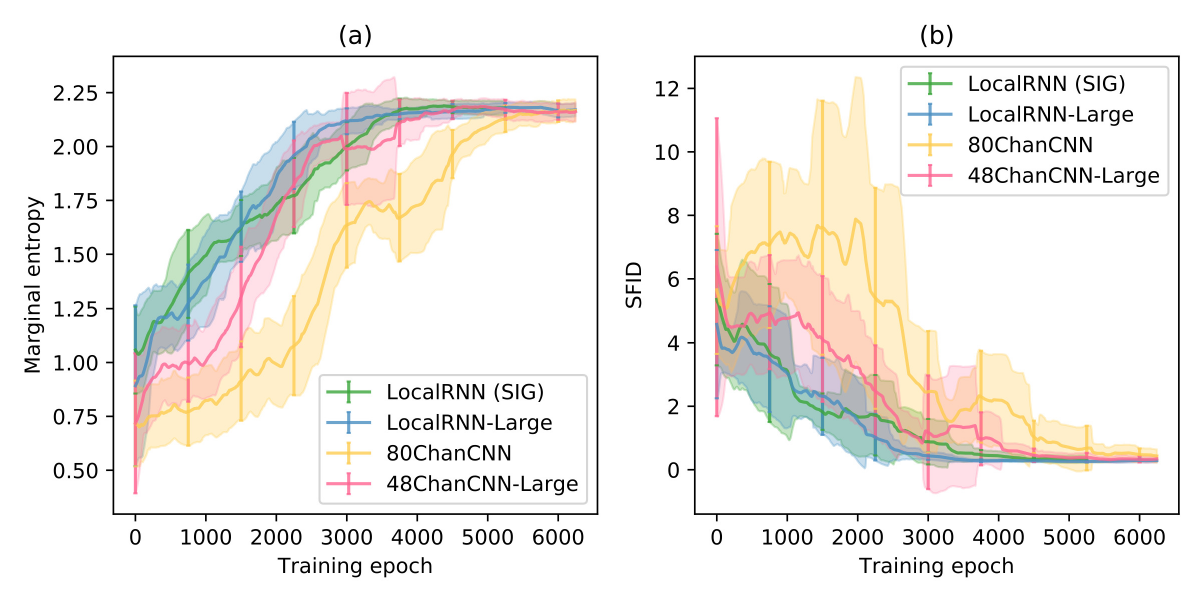}
\caption{Effects of increasing model complexity on marginal entropy (a) and SFID (b), for sequences of 40 time steps and \(\lambda_{sup} = 0 \). The large CNN architecture remains slightly behind RNN-based discriminators.}
\label{fig:is_strong_configs}
\end{figure}

Convolutional architectures are usually preferred for discriminator networks in text generation GANs as it was shown to surpass performances of RNN on a diversity of NLP tasks while being naturally suited to parallel computing~\cite{zhang2015text, gehring2017convolutional}. We therefore compare SocialInteractionGAN (interchangeably referred to as LocalRNN) with several convolutional baselines, and illustrate the results in Figure~\ref{fig:hm_cnn_rnn}. CNN discriminators were built in a similar fashion to LocalRNN, with outputs averaged from several convolutional pipelines operating at different resolutions. For comparison purposes, the same four pipelines described in section~\ref{section: exp_details} were used for convolutional discriminators, with kernels width playing the role of chunk length \(\Delta \tau\). Several variations around this standard configuration were investigated, such as increasing the number of channels or mirror padding in the dimension of actions to improve the expressive capability of the network. Models that gave the best results within the limits of our GPU memory resources were plotted in Figure~\ref{fig:hm_cnn_rnn}. Noticeably, all CNN architectures exhibit final marginal entropies close to the dataset values. In fact action sequences produced by most converged models, including LocalRNN, are hard to distinguish from real data and it is probable that the proposed task does not allow to notice differences in final performances. Nevertheless, SocialInteractionGAN consistently learns faster than the CNN baselines and exhibits a much smoother behaviour, as is particularly clear in charts (e)-(h) of Figure~\ref{fig:hm_cnn_rnn}. Differences in training speed are even stronger when supervision vanishes ((b), (d), (f) and (h)). Finally, we investigate the effects of processing action sequences at additional resolutions. We add two other pipelines to our standard configuration (i.e. two additional values of \(\Delta \tau\)), and increase the depth of convolutional blocks to the limits of our GPU capability. The resulting model is labelled as 48ChanCNN-Large. We build LocalRNN-Large in a similar fashion, augmenting the standard configuration with two additional chunk sizes. Results are shown in Figure~\ref{fig:is_strong_configs}. Although the gap between the two architectures has been partly filled, large LocalRNN still converges faster than its large convolutional counterpart. These experiments show that for the generation of discrete sequences chosen from a dictionary of human actions, it is possible to devise very efficient recurrent discriminator architectures that display more advantageous training dynamics compared to CNN discriminators, with also noticeably lighter memory footprints.

\subsubsection{Dual-stream discriminator ablation study} \label{section ablation}

We conduct an ablation study to explore the roles of the two streams of the discriminator. To that end, we run additional experiments on sequences of 40 time steps with \(\lambda_{sup}=0\), cutting off at turns the interaction and individual streams of the discriminator (respectively naming the resulting models Indiv-stream and Inter-stream). Additionally, we also compare with an architecture that has none of the two streams and that is trained only with the \(L_2\) loss (i.e.\ without adversarial loss), that we call No-GAN. The results are reported in Table~\ref{table:3}, together with the full model (Dual-stream). On the one hand, training without adversarial loss leads to much poorer scores in terms of marginal and conditional entropies, advocating in favor of the use of the adversarial loss. We hypothesize that the adversarial loss allows for a larger exploration of the action space, thus the higher marginal entropy score, and a better learning of the action sequence dynamics, as suggests the higher conditional entropy. On the other hand, disabling any of the two discriminator streams leads to degraded performances, which is especially clear in terms of SFID. In particular, we see from the high SFID that only relying on the interaction stream (Inter-stream) to produce realistic individual sequences would perform poorly. This supports our dual-stream architecture: an interaction stream alone does not guarantee sufficient individual sequence quality, but is still necessary to guide the generator into how to leverage information coming from every conversing participant.

\begin{table}
\centering
\caption{Effects of removing interaction stream (Indiv-stream), individual stream (Inter-stream), or adversarial losses altogether (No-GAN) versus the original model (Dual-stream) for sequences of 40 actions. Models are trained for 6000 epochs. Best models are those that achieve the closest \(\text{H}_\text{M}\) and \(\text{H}_\text{C}\) values to the real data and the lowest SFID.}
\label{table:3}
\resizebox{.5\textwidth}{!}{
\begin{tabular}{m{7.5em} m{5em} m{5em} m{5em}}
\toprule
Model & \(\text{H}_\text{M}\) & \(\text{H}_\text{C}\)&  SFID\\[0.5ex]
 \midrule
\textit{Real Data}&\(\textit{2.18}\)&\(\textit{0.30}\)& --\\
 \midrule
Dual-stream (SIG) & \(\textbf{2.18} \pm \textbf{0.03}\)  &  \(0.26 \pm 0.03\)  & \(\textbf{0.24} \pm \textbf{0.04}\)\\
Indiv-stream & \(2.15 \pm 0.05\)  &  \(\textbf{0.27} \pm \textbf{0.09}\)  & \(0.42 \pm 0.10\)\\
Inter-stream & \(2.23 \pm 0.03\)  &  \(0.35 \pm 0.04\)  & \(0.77 \pm 0.14\)\\
No-GAN & \(2.06 \pm 0.01\)  &  \(0.14 \pm 0.01\)  & \(0.29 \pm 0.01\)\\
 \bottomrule
\end{tabular}}
\end{table}

\subsubsection{Pushing the limits of SocialInteractionGAN} \label{section adverse}

\begin{figure}
\includegraphics[width=1\linewidth]{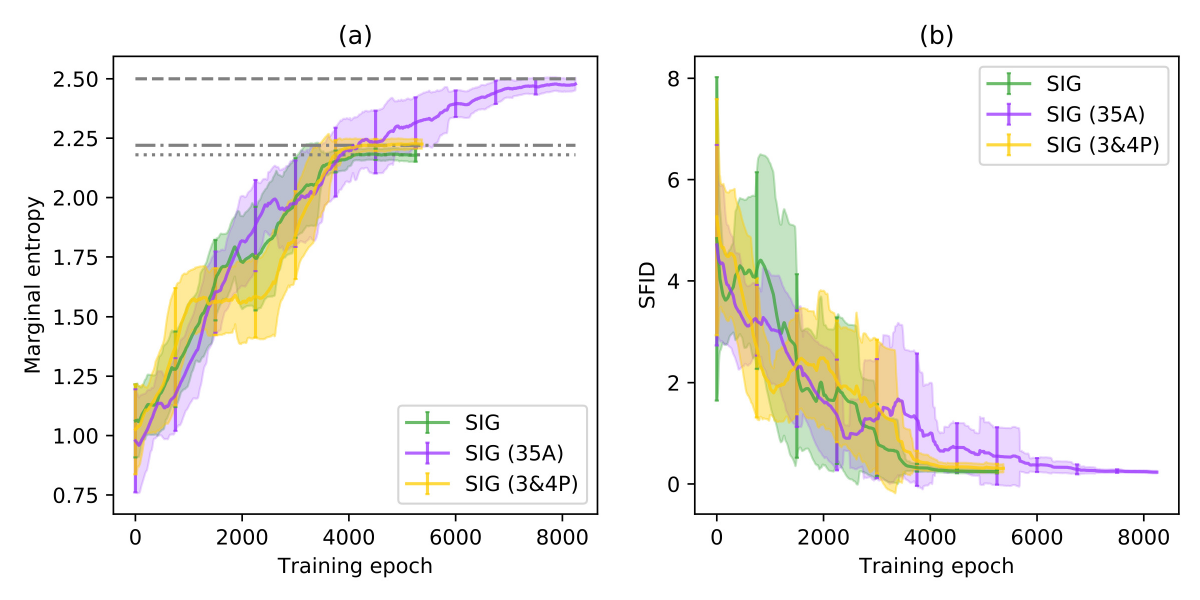}
\centering
\caption{Effects of increasing the size of action dictionary (SIG (35A)) and adding four-people interactions to the original three-people interaction dataset (SIG (3\&4P)) on marginal entropy (a) and SFID (b). Standard model is shown for comparison. All models were trained on sequences of 40 actions with \(\lambda_{sup}=0\). Gray dashed, dash-dotted and dotted lines represent respectively real data marginal entropy for large dictionary, three and four-people interactions and original SocialInteractionGAN experiments.}
\label{fig:adverse_conditions}
\end{figure}

This section illustrates the effects of enriching and diversifying the original training dataset.
Namely, we explore two variants from the initial setting: the addition of four-person interactions, and the use of a larger action dictionary. 
The first experiment consists in assessing the capacity of the network to generalize its interaction sequence predictions to larger groups of people and learn more complex action patterns, and is labeled as SIG (3\&4P).
In a second experiment, SocialInteractionGAN was trained on a larger action dictionary where we raised the cumulative occurrence threshold of dictionary entries from 90\% to 99\%.
This resulted in a dictionary containing 35 actions, more than doubling the original size, and it is therefore labeled as SIG (35A).
The marginal entropy and SFID evolution of these models for sequences of 40 actions and \(\lambda_{sup}=0\) are reported in Figure~\ref{fig:adverse_conditions}, together with the reference model (SIG).
Although SIG (3\&4P) implies training with more than 50\% additional data samples (from 600 sequences to 953), it has a very limited effect on the training dynamics. Noticeably, the model learns to match the slight increment in data marginal entropy produced by the richer interaction samples.
Regarding SIG (35A), as it could be expected from the resulting increase in network complexity, expanding the action dictionary leads to delayed convergence. However the model smoothly converges to the marginal entropy of the real data and also scores low SFID, meaning that SocialInteractionGAN is  able to handle large action dictionaries, provided a sufficient amount of training data.

\section{Conclusion}
\label{section: conclusion}
Understanding action patterns in human interactions is key to design systems that can be embedded in e.g.\ social robotic systems.
For generative models of human interactions, the main challenge resides in the necessity to preserve action consistency across participants and time.
This article presents a conditional GAN model for the generation of discrete action sequences of people involved in an interaction. To the best of our knowledge we are the first to address this problem with a data-driven solution. Although interactions are modelled in a low dimensional manifold, we believe that the principles devised here, namely the necessity to share the state between participants at every time step, and to focus the loss function on local evaluations, will apply on richer representations. We plan to extend this work on high-dimensional data such as video inputs.

The proposed model comprises an encoder-decoder generator coupled with a pooling module that allows to incorporate contextual cues to the decoder, and thus adapt a person's behaviour to the previous actions of surrounding people. We train our network with an adversarial loss, using an original dual-stream local recurrent discriminator. These two streams ensure that both generated interactions as a whole and individual action sequences that compose them must be realistic so as to minimize the adversarial loss. We present also a new calculation of IS and a Sequential FID (SFID) adapted to the quality measurement of synthetic discrete sequences, and owing to these metrics we show that our local recurrent discriminator allows for efficient training and the generation of realistic action sequences. These promising results let us foresee interesting experiments in larger dimensional discrete and continuous spaces, with possible benefits for other domains.

\bibliographystyle{IEEEtran}
\bibliography{bibliography.bib}

\begin{thebibliography}{10}
\providecommand{\url}[1]{#1}
\csname url@samestyle\endcsname
\providecommand{\newblock}{\relax}
\providecommand{\bibinfo}[2]{#2}
\providecommand{\BIBentrySTDinterwordspacing}{\spaceskip=0pt\relax}
\providecommand{\BIBentryALTinterwordstretchfactor}{4}
\providecommand{\BIBentryALTinterwordspacing}{\spaceskip=\fontdimen2\font plus
\BIBentryALTinterwordstretchfactor\fontdimen3\font minus
  \fontdimen4\font\relax}
\providecommand{\BIBforeignlanguage}[2]{{%
\expandafter\ifx\csname l@#1\endcsname\relax
\typeout{** WARNING: IEEEtran.bst: No hyphenation pattern has been}%
\typeout{** loaded for the language `#1'. Using the pattern for}%
\typeout{** the default language instead.}%
\else
\language=\csname l@#1\endcsname
\fi
#2}}
\providecommand{\BIBdecl}{\relax}
\BIBdecl

\bibitem{vinciarelli2009social}
A.~Vinciarelli, M.~Pantic, and H.~Bourlard, ``Social signal processing: Survey
  of an emerging domain,'' \emph{Image and vision computing}, vol.~27, no.~12,
  pp. 1743--1759, 2009.

\bibitem{cabrera2018matchnmingle}
L.~Cabrera-Quiros, A.~Demetriou, E.~Gedik, L.~van~der Meij, and H.~Hung, ``The
  matchnmingle dataset: a novel multi-sensor resource for the analysis of
  social interactions and group dynamics in-the-wild during free-standing
  conversations and speed dates,'' \emph{IEEE Transactions on Affective
  Computing}, 2018.

\bibitem{alameda2015salsa}
X.~Alameda-Pineda, J.~Staiano, R.~Subramanian, L.~Batrinca, E.~Ricci, B.~Lepri,
  O.~Lanz, and N.~Sebe, ``Salsa: A novel dataset for multimodal group behavior
  analysis,'' \emph{IEEE Transactions on Pattern Analysis and Machine
  Intelligence}, vol.~38, no.~8, pp. 1707--1720, 2015.

\bibitem{guo2018long}
J.~Guo, S.~Lu, H.~Cai, W.~Zhang, Y.~Yu, and J.~Wang, ``Long text generation via
  adversarial training with leaked information,'' in \emph{Proceedings of the
  AAAI Conference on Artificial Intelligence}, vol.~32, no.~1, 2018.

\bibitem{yu2017seqgan}
L.~Yu, W.~Zhang, J.~Wang, and Y.~Yu, ``Seqgan: Sequence generative adversarial
  nets with policy gradient,'' in \emph{Proceedings of the AAAI Conference on
  Artificial Intelligence}, 2017.

\bibitem{alahi2016social}
A.~Alahi, K.~Goel, V.~Ramanathan, A.~Robicquet, L.~Fei-Fei, and S.~Savarese,
  ``Social {LSTM}: Human trajectory prediction in crowded spaces,'' in
  \emph{Proceedings of the IEEE Conference on Computer Vision and Pattern
  Recognition (CVPR)}, 2016, pp. 961--971.

\bibitem{gupta2018social}
A.~Gupta, J.~Johnson, L.~Fei-Fei, S.~Savarese, and A.~Alahi, ``Social gan:
  Socially acceptable trajectories with generative adversarial networks,'' in
  \emph{Proceedings of the IEEE Conference on Computer Vision and Pattern
  Recognition (CVPR)}, 2018, pp. 2255--2264.

\bibitem{lee2017desire}
N.~Lee, W.~Choi, P.~Vernaza, C.~B. Choy, P.~H. Torr, and M.~Chandraker,
  ``Desire: Distant future prediction in dynamic scenes with interacting
  agents,'' in \emph{Proceedings of the IEEE Conference on Computer Vision and
  Pattern Recognition (CVPR)}, 2017, pp. 336--345.

\bibitem{salimans2016improved}
T.~Salimans, I.~Goodfellow, W.~Zaremba, V.~Cheung, A.~Radford, and X.~Chen,
  ``Improved techniques for training gans,'' \emph{Advances in Neural
  Information Processing Systems (NeurIPS)}, vol.~29, pp. 2234--2242, 2016.

\bibitem{heusel2017gans}
M.~Heusel, H.~Ramsauer, T.~Unterthiner, B.~Nessler, and S.~Hochreiter, ``Gans
  trained by a two time-scale update rule converge to a local nash
  equilibrium,'' \emph{Advances in Neural Information Processing Systems
  (NeurIPS)}, vol.~30, pp. 6626--6637, 2017.

\bibitem{miyato2018cgans}
T.~Miyato and M.~Koyama, ``cgans with projection discriminator,''
  \emph{International Conference on Learning Representations (ICLR)}, 2018.

\bibitem{isola2017image}
P.~Isola, J.-Y. Zhu, T.~Zhou, and A.~A. Efros, ``Image-to-image translation
  with conditional adversarial networks,'' in \emph{Proceedings of the IEEE
  Conference on Computer Vision and Pattern Recognition (CVPR)}, 2017, pp.
  1125--1134.

\bibitem{fernando2018soft+}
T.~Fernando, S.~Denman, S.~Sridharan, and C.~Fookes, ``Soft+ hardwired
  attention: An {LSTM} framework for human trajectory prediction and abnormal
  event detection,'' \emph{Neural networks}, vol. 108, pp. 466--478, 2018.

\bibitem{bartoli2018context}
F.~Bartoli, G.~Lisanti, L.~Ballan, and A.~Del~Bimbo, ``Context-aware trajectory
  prediction,'' in \emph{International Conference on Pattern Recognition
  (ICPR)}.\hskip 1em plus 0.5em minus 0.4em\relax IEEE, 2018, pp. 1941--1946.

\bibitem{zhao2019multi}
T.~Zhao, Y.~Xu, M.~Monfort, W.~Choi, C.~Baker, Y.~Zhao, Y.~Wang, and Y.~N. Wu,
  ``Multi-agent tensor fusion for contextual trajectory prediction,'' in
  \emph{Proceedings of the IEEE Conference on Computer Vision and Pattern
  Recognition (CVPR)}, 2019, pp. 12\,126--12\,134.

\bibitem{sadeghian2019sophie}
A.~Sadeghian, V.~Kosaraju, A.~Sadeghian, N.~Hirose, H.~Rezatofighi, and
  S.~Savarese, ``Sophie: An attentive gan for predicting paths compliant to
  social and physical constraints,'' in \emph{Proceedings of the IEEE
  Conference on Computer Vision and Pattern Recognition (CVPR)}, 2019, pp.
  1349--1358.

\bibitem{vasishta:hal-01875147}
P.~Vasishta, D.~Vaufreydaz, and A.~Spalanzani, ``{Building Prior Knowledge: A
  Markov Based Pedestrian Prediction Model Using Urban Environmental Data},''
  in \emph{{Proceedings of the International Conference on Control, Automation,
  Robotics and Vision (ICARCV)}}, 2018, pp. 1--12.

\bibitem{li-etal-2017-adversarial}
J.~Li, W.~Monroe, T.~Shi, S.~Jean, A.~Ritter, and D.~Jurafsky, ``Adversarial
  learning for neural dialogue generation,'' in \emph{Proceedings of the
  Conference on Empirical Methods in Natural Language Processing (CEMNLP)},
  Sep. 2017, pp. 2157--2169.

\bibitem{lin2017adversarial}
K.~Lin, D.~Li, X.~He, Z.~Zhang, and M.-T. Sun, ``Adversarial ranking for
  language generation,'' \emph{Advances in Neural Information Processing
  Systems (NeurIPS)}, vol.~30, pp. 3155--3165, 2017.

\bibitem{vezhnevets2017feudal}
A.~S. Vezhnevets, S.~Osindero, T.~Schaul, N.~Heess, M.~Jaderberg, D.~Silver,
  and K.~Kavukcuoglu, ``Feudal networks for hierarchical reinforcement
  learning,'' \emph{Proceedings of the International Conference on Machine
  Learning (ICML)}, 2017.

\bibitem{hochreiter1997long}
S.~Hochreiter and J.~Schmidhuber, ``Long short-term memory,'' \emph{Neural
  computation}, vol.~9, no.~8, pp. 1735--1780, 1997.

\bibitem{bahdanau2014neural}
D.~Bahdanau, K.~Cho, and Y.~Bengio, ``Neural machine translation by jointly
  learning to align and translate,'' \emph{International Conference on Learning
  Representations (ICLR)}, 2015.

\bibitem{zhang2015text}
X.~Zhang and Y.~LeCun, ``Text understanding from scratch,'' \emph{arXiv
  preprint arXiv:1502.01710}, 2015.

\bibitem{gehring2017convolutional}
J.~Gehring, M.~Auli, D.~Grangier, D.~Yarats, and Y.~N. Dauphin, ``Convolutional
  sequence to sequence learning,'' \emph{Proceedings of the International
  Conference on Machine Learning (ICML)}, pp. 1243--1252, 2017.

\bibitem{lim2017geometric}
J.~H. Lim and J.~C. Ye, ``Geometric gan,'' \emph{arXiv preprint
  arXiv:1705.02894}, 2017.

\bibitem{szegedy2016rethinking}
C.~Szegedy, V.~Vanhoucke, S.~Ioffe, J.~Shlens, and Z.~Wojna, ``Rethinking the
  inception architecture for computer vision,'' in \emph{Proceedings of the
  IEEE Conference on Computer Vision and Pattern Recognition (CVPR)}, 2016, pp.
  2818--2826.

\bibitem{mccowan2005ami}
I.~McCowan, J.~Carletta, W.~Kraaij, S.~Ashby, S.~Bourban, M.~Flynn,
  M.~Guillemot, T.~Hain, J.~Kadlec, V.~Karaiskos \emph{et~al.}, ``The ami
  meeting corpus,'' in \emph{Proceedings of the 5th international conference on
  methods and techniques in behavioral research}, vol.~88.\hskip 1em plus 0.5em
  minus 0.4em\relax Citeseer, 2005, p. 100.

\bibitem{KOUTSOMBOGERA18.596}
M.~Koutsombogera and C.~Vogel, ``\BIBforeignlanguage{english}{Modeling
  collaborative multimodal behavior in group dialogues: The multisimo
  corpus},'' in \emph{\BIBforeignlanguage{english}{Proceedings of the Eleventh
  International Conference on Language Resources and Evaluation (LREC
  2018)}}.\hskip 1em plus 0.5em minus 0.4em\relax Paris, France: European
  Language Resources Association (ELRA), may 2018.

\bibitem{kossaifi2019sewa}
J.~Kossaifi, R.~Walecki, Y.~Panagakis, J.~Shen, M.~Schmitt, F.~Ringeval,
  J.~Han, V.~Pandit, A.~Toisoul, B.~Schuller \emph{et~al.}, ``Sewa db: A rich
  database for audio-visual emotion and sentiment research in the wild,''
  \emph{IEEE transactions on pattern analysis and machine intelligence},
  vol.~43, no.~3, pp. 1022--1040, 2019.

\bibitem{ba2016layer}
J.~L. Ba, J.~R. Kiros, and G.~E. Hinton, ``Layer normalization,'' \emph{CoRR},
  vol. abs/1607.06450, 2016.

\bibitem{miyato2018spectral}
T.~Miyato, T.~Kataoka, M.~Koyama, and Y.~Yoshida, ``Spectral normalization for
  generative adversarial networks,'' \emph{International Conference on Learning
  Representations (ICLR)}, 2018.

\bibitem{ioffe2015batch}
S.~Ioffe and C.~Szegedy, ``Batch normalization: Accelerating deep network
  training by reducing internal covariate shift,'' in \emph{International
  Conference on Machine Learning (ICML)}, 2015, pp. 448--456.

\bibitem{Kingma2015AdamAM}
D.~P. Kingma and J.~Ba, ``Adam: A method for stochastic optimization,''
  \emph{CoRR}, vol. abs/1412.6980, 2015.

\end{thebibliography}
\balance

\end{document}